\documentclass{article}

\usepackage{arxiv}

\usepackage[utf8]{inputenc} 
\usepackage[T1]{fontenc}    
\usepackage{hyperref}       
\usepackage{url}            
\usepackage{booktabs}       
\usepackage{amsfonts}       
\usepackage{nicefrac}       
\usepackage{microtype}      
\usepackage{lipsum}

\usepackage{algorithmic}
\usepackage[]{algorithm2e}
\usepackage{subcaption}
\usepackage{amsmath}
\usepackage{makeidx}
\usepackage{amssymb}
\usepackage{graphicx}
\usepackage{bbm}

\title{A Review of Reinforcement Learning for Autonomous Building Energy Management}

\author{
  Karl Mason\\
  School of Electrical and Computer Engineering\\ 
  Georgia Institute of Technology\\ 
  Atlanta, GA, USA\\
  \texttt{kmason35@gatech.edu} \\
   \And
 Santiago Grijalva \\
  School of Electrical and Computer Engineering\\ 
  Georgia Institute of Technology\\ 
  Atlanta, GA, USA\\
  \texttt{sgrijalva@ece.gatech.edu} \\
}

\begin{document}
\maketitle 

\begin{abstract}
The area of building energy management has received a significant amount of interest in recent years. This area is concerned with combining advancements in sensor technologies, communications and advanced control algorithms to optimize energy utilization. Reinforcement learning is one of the most prominent machine learning algorithms used for control problems and has had many successful applications in the area of building energy management. This research gives a comprehensive review of the literature relating to the application of reinforcement learning to developing autonomous building energy management systems. The main direction for future research and challenges in reinforcement learning are also outlined.
\end{abstract}



\section{Introduction}\label{sec:Intro}

Meeting the global energy needs in an efficient and sustainable manner is one of the most pressing issues facing society today. The energy consumption from buildings accounts for approximately 40\% of the total global energy consumption \cite{nejat2015global}. It is therefore imperative to find innovative ways to help reduce and optimize the energy consumed by buildings. Some of the main sources of energy consumption in buildings include: Heating Ventilation and Air Conditioning (HVAC), water heating, and lighting. Reducing the energy consumed by buildings has many benefits. The consumer is benefited with lower energy bills. Electricity providers benefit from reduced peak loads that must be met. There is also a wider benefit to society as a whole from reducing building energy consumption and reduction of emissions. Much of the energy that powers the grid comes from greenhouse gas emitting sources, e.g. coal. Therefore, a reduction in building energy consumption would also correspond to a reduction in the emission of greenhouse gasses such as carbon dioxide.

The prevalence of digital systems in every aspect of modern society is largely due to the steady reduction in the cost and size of micro-processors, along with their increased computing power. The widespread availability of computing has manifested itself in the area of building energy consumption in the form of advanced sensors and actuators, and building energy management systems \cite{zhou2016smart}. These systems consist of monitoring and controlling the activity of each device within the building with the aim of reducing the demand placed onto the grid and the cost to the consumer. The difficulty of this management process is increased with the addition of photovoltaic (PV) panels, batteries, electric vehicles, and smart appliances. Building energy management systems are comprised of many components. Sensors are needed to monitor the building, e.g. temperature, humidity and load sensors. Communication is needed between devices in order to facilitate the monitoring and scheduling of their operation. This communication between devices is an example of the ``Internet of Things'' (IoT). Smart meters are needed to record energy consumption and to communicate with energy providers. Many devices are equipped with rudimentary control mechanisms such as rule-based decision systems. For example, the thermostat that controls a HVAC system turns heating and cooling on/off when the temperature of the room crosses a user-defined threshold. This paper will provide an overview of how more sophisticated machine learning control systems, in particular Reinforcement Learning (RL), have improved the operation of building energy management systems.

Reinforcement Learning (RL) is a form of machine learning that consists of an agent interacting in an environment, learning what actions to take depending on the state of the environment \cite{sutton2018reinforcement}. The agent learns by trial and error and is rewarded for taking desirable actions. The environment is often modelled as a Markov Decision Process (MDP). RL algorithms date back to the 1970s and 1980s. One of the more popular RL algorithms, Q Learning, was first proposed in 1989 \cite{Watkins89}. These algorithms have been applied to a wide range of problems over the years, everything from traffic light control \cite{mannion2016experimental} to watershed management \cite{mason2016applying}. The combination of RL algorithms with deep neural networks has significantly increased the effectiveness of RL methods and has enabled them to be applied to tasks involving computer visions, e.g. self-driving cars. Recent notable achievements of RL have resulted in a significant amount of research into RL and its applications. These achievements include mastering the game of Go and playing Atari games.

Many researchers have since begun applying reinforcement learning to some of the challenging areas of building energy management. RL has been applied to tasks such as HVAC control, water heater control, electric vehicle charging, lighting control and appliance scheduling. An advantage of applying RL to address these problems is that the algorithm learns by itself what the best control policy is. When implementing more traditional rule-based approaches, the designer must handcraft the thresholds that the system will adhere to. This na\"{i}ve approach is not necessarily able to minimize energy consumption to the extent that RL can. There are several factors that increase the complexity of applying RL to these problems, such as identifying what state information is needed, conflicting objectives and simulator design. These will be discussed in detail throughout the paper. The main contributions of this research are:

\begin{enumerate}
    \item To give a comprehensive review of the literature relating to the application of RL to building energy management.
    
    \item To quantify the impact of RL on building energy management systems in terms of energy savings.
    
    \item To establish the limitations of RL for building energy management and outline potential areas of future research.
    
\end{enumerate}

The outline of the paper is as follows. Sections 2 will give outline the research area of building energy management. An overview of reinforcement learning will be provided in Section 3. Section 4 will give a comprehensive account of the applications of reinforcement learning within the area of building energy management. Section 5 will discuss some of the limitations and potential areas for future work concerning the application of reinforcement learning to building energy management. Finally, Section 6 will outline what can be concluded as a result of this research.

\section{Autonomous Building Energy Management}
\label{sec:smartBuildings}
The development of autonomous building energy management systems has received a lot of interest in recent years for a number of reasons, one of the primary reasons being the drive to increase the energy efficiency in buildings through operational methods. As stated in the introduction, energy consumption in buildings accounts for a significant portion of the total worldwide energy consumption \cite{nejat2015global}. There are a number of factors that influence how energy efficient a building is, e.g. insulation, construction materials and overall design. The aim of developing autonomous building energy management systems is to further reduce both overall energy consumption and the cost of powering buildings by utilizing advancements in technologies such as:

\begin{itemize}
    \item Autonomous control systems

    \item Internet of Things (IoT)
    
    \item Batteries
    
    \item Smart grid 
\end{itemize}

\subsection{HVAC}
\label{sec:HVAC}

Heating Ventilation and Air Conditioning (HVAC) is one of the most energy intensive consumers of energy in buildings. It is estimated that HVAC is responsible for 50\% of building energy consumption in the US and between 10 - 20\% of overall consumption in developed countries \cite{perez2008review}. This is due to expectation of thermal comfort in developed countries, rather than it being regarded as a luxury \cite{yang2014thermal}. It is also well known that there is a strong correlation between outside temperatures and HVAC energy consumption \cite{li2012impact}. This energy consumption is expected to increase due to the increased number of extreme weather events observed globally \cite{coumou2012decade}. The most basic HVAC control system would be a threshold-based approach in which a thermostat is used to regulate the temperature. If the temperature crosses some predefined threshold, the HVAC system activates to heat/cool the environment as required. There have been a wide number of strategies proposed to control the operation of HVAC systems \cite{afram2014theory}. Section \ref{sec:BuildingManageRL} will outline previous applications of RL to HVAC control.

\subsection{Lighting and Appliances}
\label{sec:LightAppliances}

There are of course many other energy intensive devices within the buildings besides HVAC. Much of the energy consumed in domestic dwellings comes from: lighting, televisions, water heaters, washing machines, dryers, fridge freezers, etc. In order to reduce energy consumption of appliances, the design of these devices has been significantly improved in recent years. For example, some dryers are designed to have moisture sensors to prevent over drying. A significant amount of effort has focused on changing consumer behaviour, e.g. turning lights off when the room is not in use or turning off television sets when not in use. It is known that consumer behaviour plays a significant role in energy consumption \cite{gram2013efficient}. Previous studies have shown that providing regular feedback to consumers of their energy consumption can reduce their consumption by 15\% on average \cite{wood2003dynamic}.  Another crucial issue with regards to the energy consumption of appliances is the variation in energy demand throughout the day. This affects the load demand placed on the electrical grid. This will be discussed further in Section \ref{sec:grid}. This provides an opportunity for autonomous building energy management systems to help reduce these peak demands by managing energy consumption, e.g. dimming lights at peak times. Autonomous control systems are now being developed to effectively schedule appliances to minimize energy cost \cite{adika2014autonomous}.


\subsection{Electric Vehicles and Batteries}
\label{sec:EV_Batteries}
The automotive industry is also heavily influencing the behaviour of residential energy consumption due to the current transition from fossil fuel powered vehicles to electric vehicles. The need to draw power from the grid in order to charge electric vehicles significantly increases the electrical load placed on the grid. Grahn et al. investigated the affect Plug in Hybrid Electric Vehicles (PHEVs) have on the domestic energy consumption in contrast to other appliances in the home \cite{grahn2013phev}. The authors of this study estimate that if all of the 4.3 million vehicles in Sweden were electric, this would correspond to an additional 34 GWh of electricity consumption. This is approximately 10 \% of Sweden's daily electricity consumption \cite{grahn2013phev}. This is a significant additional load for the grid to adapt to. 


Vehicle to grid (V2G) technology is a promising solution to address this problem \cite{tan2016integration}. Vehicle to grid technology works by selling electricity back to the grid when the car is not being used. A variation V2G is to consider the electric vehicle as a deferrable load. When the grid needs more power, charging of electric vehicle can stop for a few seconds or minutes. In this process there is no actual injection of power from the vehicle to the grid, but a pause in charging that can help the grid. V2G also has the potential to alleviate some of the load on the grid during peak demand. This further motivates the need for effective building energy management systems. For example, a potentially effective strategy would be to charge the vehicle overnight when electricity is cheapest and to sell energy back to the grid at 7 pm when demand is high. Electric vehicles could also contribute to address the problem of incorporating variable renewable energy into the grid. The variability of renewable sources is a challenge to incorporating technologies such as Photovoltaic (PV) into the grid \cite{kempton2005vehicle}. This will be discussed further in Section \ref{sec:grid}. Battery technology in general is expected to play a major role in addressing some of the issues raised above \cite{dunn2011electrical}. The Tesla Powerwall is a prime example of this \cite{truong2016economics}. Energy management systems have already been developed that utilize battery technology to help manage the electrical demand \cite{tani2015energy}.

\subsection{Smart Meters and the Internet of Things}
\label{sec:smartMeters}

Smart meters are a key development that will enable the large-scale deployment of autonomous building energy management systems \cite{depuru2011smart}. They enable communication with electricity providers that aids their operation. Smart meters also enable the control of appliances within the home. Smart meters are already in widespread use. Italy and Sweden were two of the first to countries to have completed their deployment of smart meters nation-wide, with many other countries making significant progress in their deployment \cite{uribe2016state}.

The term ``Internet of Things'' or IoT, refers to a network of devices, appliances, sensors and electronics that can connect with one another \cite{stojkoska2017review}. It is this communication between devices that enables building energy management systems to operate.

\subsection{Electrical Grid and Renewables}
\label{sec:grid}

One of the main challenges when developing autonomous building control systems is integrating it with the electrical grid. These systems need to schedule their operation in order to minimize the cost of electricity to the consumer. This is further complicated with many individuals installing PV solar panels, as previously discussed. This gives rise to the recent paradigm of ``prosumers'' (producers and consumers) \cite{grijalva2011prosumer}. Within the prosumer paradigm, individuals both consume energy from the grid and produce energy which is sold to the grid. This creates issues of instability within the grid which must be addressed \cite{nazari2014distributed}. This is due to feeders with high levels of PV penetration producing an excess of electricity that the grid struggles to cope with \cite{reno2013smart}. 

Another factor which increases the complexity of implementing autonomous building control systems is the price of energy \cite{hu2013hardware}. Electricity prices vary based on demand and are therefore considered by many control systems \cite{oldewurtel2010reducing}. Weather also has a significant effect on energy consumption. As stated previously, energy consumption via HVAC increases significantly with both exceedingly high and low outside weather temperatures \cite{li2012impact}. In addition to this, sunny weather results in a higher output from PV systems, this further exacerbates the problems mentioned relating to grid instabilities \cite{nazari2014distributed}.  

The next section will outline Reinforcement learning, a popular machine learning method that has been applied to many problems relating to smart grid, building energy management, and smart homes.

\section{Reinforcement Learning}
\label{sec:RL}

\begin{figure}[h]
    \centering
    \includegraphics[width=0.6\textwidth]{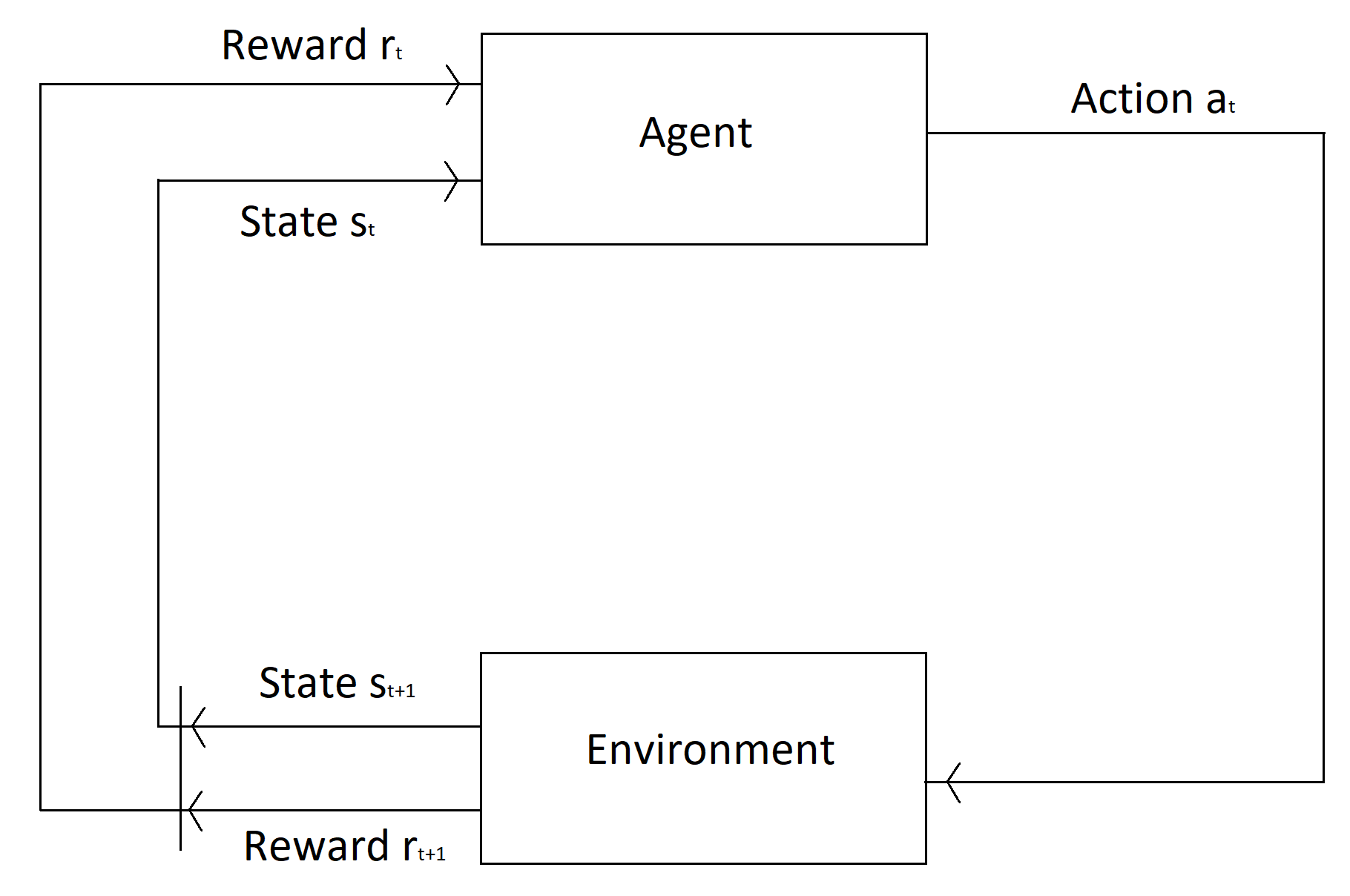}
    \caption{Agent Environment Interaction}
    \label{fig:RLAgentEnvironment}
\end{figure}

Reinforcement Learning (RL) is a subgroup of machine learning research that involves an agent learning by itself what actions to take in an environment so that it maximizes some reward \cite{sutton2018reinforcement}. This typically involves a significant amount of trial and error from the agent as it learns what actions result in the highest reward. Figure \ref{fig:RLAgentEnvironment} illustrates this interaction with its environment. Algorithm \ref{Alg:RL} presents a generic pseudocode, outlining the broad steps taken in a typical RL algorithm. The agent interacts with its environment in discrete time steps. An agent typically has a policy ($\pi$) which determines what actions it will take. The goal is then to find the optimum policy $\pi\textsuperscript{*} \in \Pi$, where $\Pi$ is the set of possible policies. The value of a policy $\pi$ in a given state s is calculated using the value function in Equation \ref{eqn:ValFunct}.

\begin{equation}
  V^{\pi}(s) = E[\Sigma_{k=0}^{\infty} \gamma^k r_{t+k+1} \mid s, \pi]
  \label{eqn:ValFunct}
\end{equation}

Where E is the expected future return and $\gamma$ is the discount factor. In order for the agent to assess the value of its current state, it must also consider the expected future rewards it will obtain if it follows its current policy. Equation \ref{eqn:ValFunct} quantifies this mathematically. The discount factor $\gamma$ determines how much weighting the agent gives to future rewards, as discussed in Section \ref{sec:Problem}. The value function of state s for the optimum policy $\pi^{*}$ is stated in Equation \ref{eqn:ValFunctOpt}, calculated using the Bellman equation.

\begin{equation}
  V^{\pi^{*}}(s) = max E[r_{t+1} + \gamma V^{\pi^{*}}(s_{t+1}) \mid s, \pi^{*}]
  \label{eqn:ValFunctOpt}
\end{equation}

\begin{algorithm}[]
\caption{Generic Reinforcement Learning Algorithm}
\begin {algorithmic}
\STATE Initialize Agent\\
\While{Episode e \textless Emax}{
    \STATE Initialize Environment\\
    \For{Time $t = 1$ \textbf{to} T}{
        Observe state $s_t$ \\
        Select action a $\in A$ \\
        Observe reward r and new state s$'$ \\
        Update policy $\pi$ \\
        Transition to new state $s_{t+1}$\\
    }
}
\end{algorithmic}
\label{Alg:RL}
\end{algorithm}

One of the first significant achievements of RL was its application to learn to play backgammon in 1995 by Gerald Tesauro \cite{tesauro1995temporal}. Some of the more recent achievements include learning to play Atari games at a level comparable to an expert human player \cite{mnih2015human} in 2015 and mastering the game of GO \cite{silver2016mastering} in 2016 by beating the world champion Lee Sedol. This is by no means a comprehensive list of the many achievements of RL algorithms. RL has been successfully applied to a vast range of problems including building energy management, which will be discussed later.

\subsection{Problem Types}
\label{sec:Problem}

The most common way to model an RL problem is as a Markov Decision Process (MDP) \cite{sutton2018reinforcement}. A MDP is a discrete time framework for modelling decision making. A MDP is defined as a tuple $\langle S, A, T , R, \gamma \rangle $, where $S$ is the set of states the environment can be, A is the set of possible actions, $T = Pr_{a}(s,s')= Pr(s_{t+1} = s' \mid s_t = s, a_t = a)$ is the probability that taking action a in state s will lead to state s$'$ at time t+1, and finally $R = R_{a}(s,s')$ is the reward received after transitioning from state s to s$'$. RL agents account for future rewards using a discount factor $\gamma \in [0,1]$, a higher value makes the agent more forward thinking. 

An extension to the standard MDP is the Partially Observable Markov Decision Process (POMDP) \cite{spaan2012partially}. In POMDPs it is assumed that the underlying problem is a MDP, however the agent does not have complete knowledge of the state of the environment. A POMDP is defined as the tuple $\langle S, A, T , R, \Omega, O, \gamma \rangle $. Here $\Omega $ and $ O$ are the set of observations and the set of conditional observation probabilities respectively. When the environment transitions to state s$'$, the agent receives the observation $o \in \Omega$ with probability $O(o \mid s', a)$.

For problems that require multiple agents, RL is often combined with Multi-Agent Systems (MAS) \cite{wooldridge2009introduction}. This is referred to as Multi-Agent Reinforcement Learning (MARL) \cite{busoniu2010multi}. One of the key challenges of MARL is to ensure that agents coordinate their actions to achieve a global optimal result. This can be difficult to achieve in practice as each agent may only have partial information of the environment, making it a decentralized problem. These sorts of problems can be modeled as a Decentralized Partially Observable Markov Decision Process (Dec-POMDP) \cite{amato2013decentralized}. Dec-POMDPs are defined by the tuple $\langle I, S, \textbf{A}, T , R, \boldsymbol{\Omega}, O, h \rangle $. In this case, $I$ is the set of agents, $\textbf{A} = \times A_i$ is the joint action space, $\boldsymbol{\Omega} = \times \Omega_i$ is the joint observation space and finally h is the horizon. 

Another type of RL problem domain are problems that consist of multiple tasks, i.e. Multi Task Learning (MTL). In this type of problem, each task $T_j$ has a tuple $\langle D, T_j, R_j, O_j\rangle$ where $D$ is the underlying domain, $T_j$ is the task specific transition, $R_j$ is the task specific reward and $O_j$ is the task specific observation. This is closely related to Transfer Learning (TL), in which the goal is to transfer knowledge learned when solving one problem to solve a different problem \cite{taylor2009transfer}. The distinction being that in MTL, the tasks are variants of the same underlying problem.

Many RL problems have multiple objectives. These problems are referred to as Multi-Objective RL (MORL) problems \cite{van2014multi}. In MORL, the MDP is extended to be a Multi-Objective MDP (MOMDP) where the agent receives a vector reward signal $\textbf{R}(s,a) = (R_1(s,a), R_2(s,a), ..., R_m(s,a))$ where m is the number of objectives. The goal here is to learn multiple policies simultaneously for each objective \cite{mannion2017policy,mannion2016dynamic}.

The task of learning, itself can be viewed as a learning problem. This is referred to as Meta learning \cite{lemke2015metalearning}. Meta learning has been applied to RL algorithms. For example the RL\textsuperscript{2} algorithm consists of a fast and slow RL algorithm \cite{duan2016rl}. The fast RL learner learns how to complete the task while the slow RL learner learns the optimum parameters of the fast learner.

Multi-Agent RL problems can also be either cooperative or competitive \cite{tampuu2017multiagent}. In cooperative MARL, each agents' objective aligns with one another either in the form of a shared goal or goals that do not conflict with one another. In competitive environments, agents have conflicting objectives whereby a gain for one agent results in a loss for another.

\subsection{Model-Free vs Model-Based}
\label{sec:ModelUse}
RL algorithms can be subcategorized into two groups, model-based and model-free algorithms. In model-based algorithms, the agent learns a model of the environment based on its observing how the state of the environment changes when certain actions are taken. These observations are used to estimate the environments state transition function $T(s' \mid s, a)$ and reward $R$. Once the algorithm learns a model of the environment, it can be combined with a planning algorithm to decide what actions to take \cite{ray2010model}. Examples of model-based algorithms include: Explicit Explore and Exploit (E \textsuperscript{3}) \cite{kearns2002near}, Prioritized sweeping \cite{moore1993prioritized}, Dyna \cite{sutton1991dyna} and Queue-Dyna \cite{peng1993efficient}.

Model-free approaches do not need to develop a model of the environment. Instead model-free approaches learn a policy, via trial and error, with the aim of approximating the optimum policy. Many of the routinely used RL algorithms are model-free, e.g. Q Learning \cite{Watkins89} and SARSA \cite{rummery1994line}. Model-free approaches are more popular in the literature as they are generally less computationally expensive. Model-based approaches first require building an accurate model of the problem, which can be difficult. Once an accurate model of the environment is obtained, the algorithm must find the optimum policy by planning ahead.

\subsection{Discrete and Continuous Search Spaces}
\label{sec:StateSpaces}

\begin{figure}[h]
    \centering
    \includegraphics[width=0.6\textwidth]{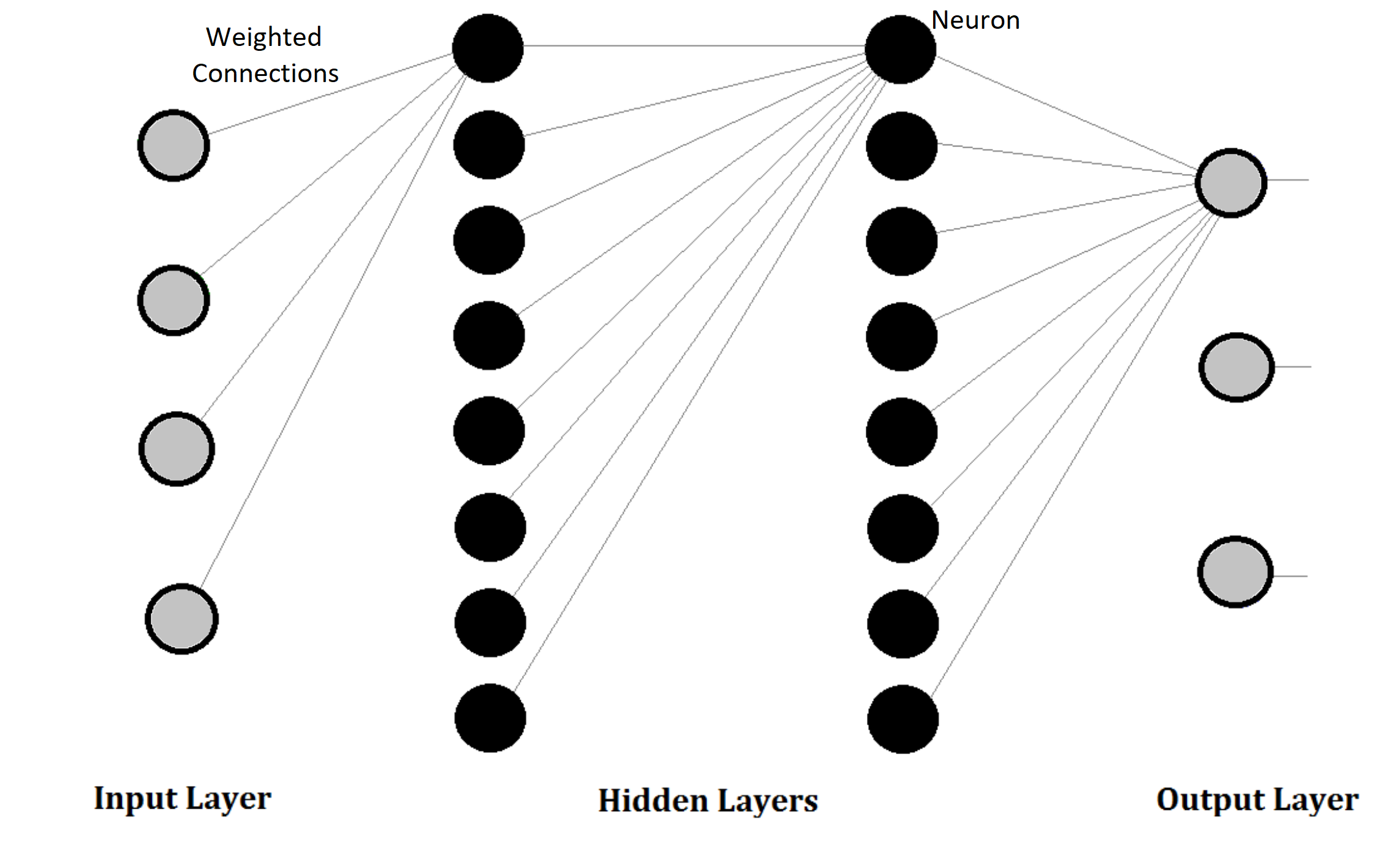}
    \caption{Neural Network Function Approximator \cite{mason2018advances}}
    \label{fig:NN}
\end{figure}

Many RL algorithms, such as Q learning \cite{Watkins89}, represent the values of each state action pair in a look up table (a Q table for Q learning). This approach works well for problems where the state action space is small. However, a fundamental limitation with the tabular approach for representing Q values is its scalability. It's clear that when the number of states and actions increase, the size of the table quickly becomes exceedingly large. The state action space becomes infinite for problems with continuous variables. The widely adopted solution to this is to replace the Q table with a function approximator, most commonly a neural network. When implementing a neural network to represent the Q function, the network reads in the state of the environment as input. The output of the network is the Q value for each action. This function approximation allows RL algorithms to deal with exceedingly large state spaces such as those with images. This is the approach taken for all of the recent work in deep RL and Atari games \cite{mnih2015human}. The image in Figure \ref{fig:NN} illustrates a deep neural network that can be used as a function approximator for RL algorithms.

Function approximation works well for problems with continuous state spaces. This approach still presents a problem for domains with continuous action spaces. One approach would be to discretize the action space, however this will limit the performance of the RL algorithm. A number of RL algorithms have been proposed to address problems with continuous action variables, including: Deep Deterministic Policy Gradient (DDPG) \cite{lillicrap2015continuous} and Sequential Monte Carlo (SMC) \cite{lazaric2008reinforcement}.

\subsection{Q Learning}
\label{sec:QLearn}

One of the most prominent RL algorithms is Q Learning \cite{Watkins89}. The Q Learning algorithm has the properties of being an off-policy and model-free reinforcement learning method. Off-policy refers to an agent that learns the value of its policy independent of its actions \cite{Watkins92}. When applied to discrete state action spaces, the agents policy is determined by its Q table, a matrix where $Q = {S \times A}$. These Q values are used to select what action to take in a given state. Once the agent takes an action (a) for the current environmental state (s), the agent receives a reward (r). The agent then updates it's Q table based on Equation \ref{eqn:Q-Learning}.

\begin{equation}
  Q(s_t,a_t)\leftarrow Q(s_t,a_t)+\alpha[r+\gamma \max_{a_{t}}Q(s_{t+1},a_{t})-Q(s_t,a_t)]
  \label{eqn:Q-Learning}
\end{equation}

Where $\alpha \in [0,1]$ is the learning rate and $\gamma \in [0,1]$ is the discount factor. It is also common to implement Q Learning with an $\epsilon$-greedy selection policy where a random action is selected with probability $\epsilon \in [0,1]$. The purpose of this is to encourage exploration. Another option is to use a softmax action selection policy, outlined in Section \ref{sec:Actor-Critic}.

\subsection{SARSA}
\label{sec:SARSA}

Another popular RL algorithm is SARSA (State Action Reward State Action) \cite{rummery1994line}. As is the case with Q Learning, SARSA is a model-free learning algorithm. Unlike Q Learning, SARSA is an on-policy learning approach meaning that it estimates the value of its policy at a given time while using that particular policy. SARSA updates its Q values using Equation \ref{eqn:SARSA}.

\begin{equation}
 Q(s_t,a_t)\leftarrow Q(s_t,a_t)+\alpha[r+\gamma Q(s_{t+1},a_{t+1})-Q(s_t,a_t)]
  \label{eqn:SARSA}
\end{equation}

Where each parameter has the same meaning as in Equation \ref{eqn:Q-Learning}.

\subsection{Actor Critic}
\label{sec:Actor-Critic}

Actor Critic (AC) is a RL algorithm that consists of two agents: an actor and a critic. The actor makes decisions based on its observations of the environment and current policy. The purpose of the critic is to observe both the state of the environment and the reward obtained from the environment based on the actor's decision. The critic then gives feedback to the actor \cite{gajjar2003application}. Actor critic is an on-policy and model-free RL algorithm.

In actor critic learning, the actor takes an action a. This results in a Temporal Difference (TD) error, calculated using Equation \ref{eqn:AC_TDerror}.
\begin{equation}
  \delta_t = r_{t+1} + \gamma V(s_{t+1}) - V(s_t)
  \label{eqn:AC_TDerror}
\end{equation}

Where V is the current estimate of the state value. The critic then updates its estimate of the value function using Equation \ref{eqn:AC_Vupdate}.
\begin{equation}
  V(s_t) \leftarrow V(s_t) + \alpha \delta_t
  \label{eqn:AC_Vupdate}
\end{equation}

Where $\alpha$ is the learning rate. The actor also uses the TD error to update its preference (p) to select that action again via Equation \ref{eqn:AC_Prefupdate}
\begin{equation}
  p(s_{t}, a_{t}) \leftarrow p(s_{t}, a_{t}) + \beta \delta _{t}
  \label{eqn:AC_Prefupdate}
\end{equation}

Where $\beta \in [0,1]$. The actor then uses this preference in combination with a softmax selection method to choose its action. Its policy is outlined in Equation \ref{eqn:AC_policy}.

\begin{equation}
  \pi(s_t, a_t) = \frac{exp (\frac{Q(s, a)}{\tau})}{\Sigma^{B}_{b=1} exp(\frac{Q(s, a)}{\tau})}
  \label{eqn:AC_policy}
\end{equation}

\begin{figure}[h]
    \centering
    \includegraphics[width=0.5\textwidth]{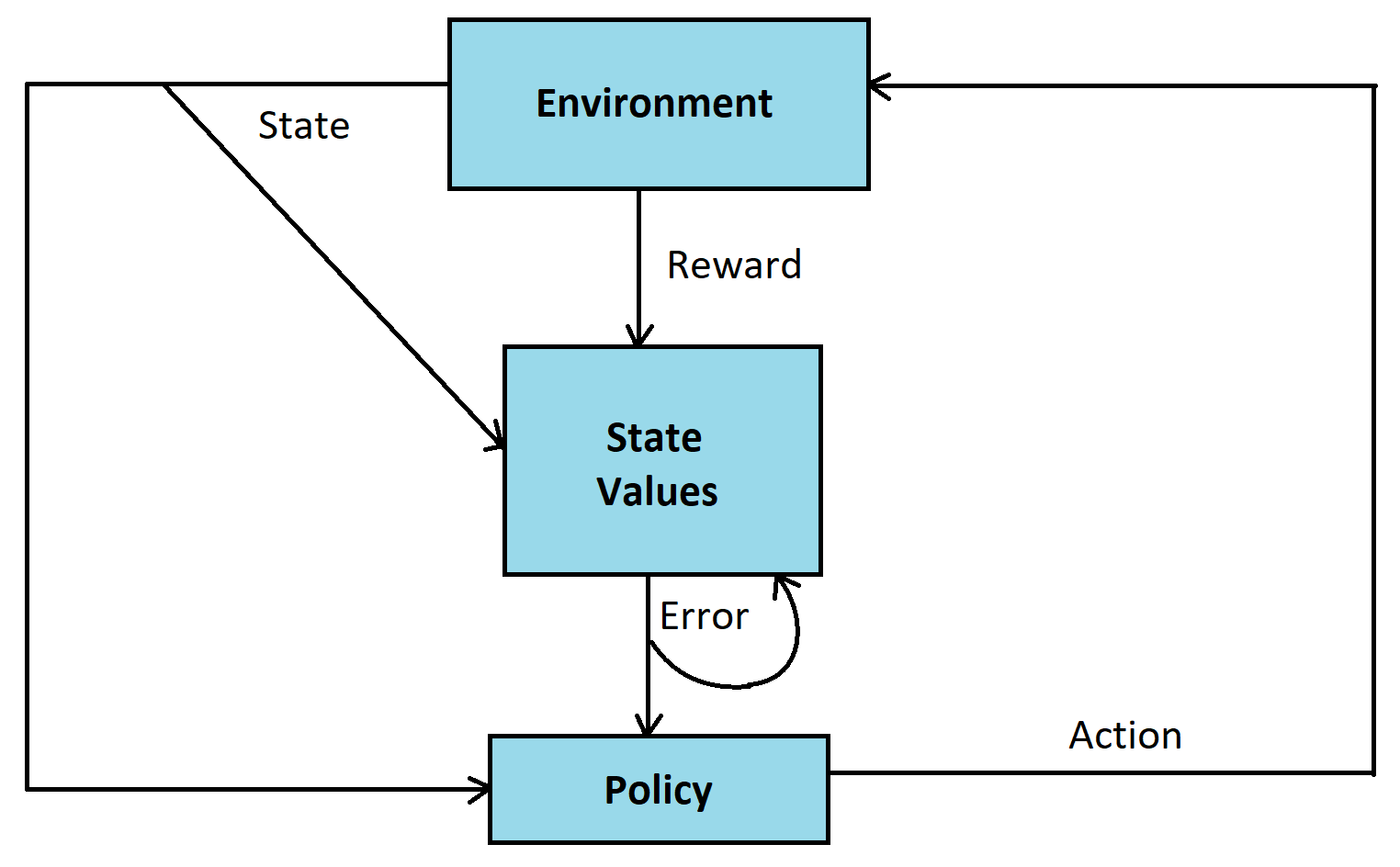}
    \caption{Actor Critic Environment Interaction}
    \label{fig:ACEnv}
\end{figure}

Where $\tau$ is the temperature and is used to calibrate the Softmax selection policy. Lower $\tau$ values increase the likelihood of selecting the best estimated action while higher values make each action selection equally likely. As stated in Section \ref{sec:QLearn}, $\epsilon$-greedy is another popular action selection policy. While Figure \ref{fig:RLAgentEnvironment} illustrates how an agent generally interacts with its environment, Figure \ref{fig:ACEnv} specifically illustrates how AC agents interact with their environment.

\subsection{Recent Methods}
\label{sec:RecRL}

The algorithms outlined in the previous sections are still routinely used in the literature. There have been some significant advances made in RL that have led to new and powerful techniques. The most significant of these being the combination of RL with deep neural networks as previously mentioned in Section \ref{sec:StateSpaces}. One of the more popular recent methods is the Deep Q Network (DQN) algorithm \cite{mnih2015human}. DQN utilizes both an experience replay of past state action pairs to evaluate policies and a periodic update for the target network. The parameters ($\theta$) of the policy function (Q) are updated by calculating the error between the policy network (Q) and the target network (\^Q) parameterised by $\theta^{-}$, as shown in Equation \ref{eqn:DQN_update}.

\begin{equation}
  L_i (\theta_i)= \mathbb{E}_{(s,a,r,s') \sim U (D)} (r_i + \gamma max_{a'} \hat{Q}(s', a' ; \theta^{-}_i) - Q(s,a;\theta_i))^2
  \label{eqn:DQN_update}
\end{equation}

Where $\mathbb{E}_{(s,a,r,s') \sim U (D)}$ is an episode $(s,a,r,s')$ sampled uniformly from the dataset D of previous experiences. The DQN algorithm was applied to a range of Atari games and found to exceed human level performance.

Another recent RL algorithm is the Asynchronous Advantage Actor Critic (A3C) algorithm \cite{mnih2016asynchronous}. As mentioned in Section \ref{sec:Actor-Critic}, actor critic methods consists of a policy $\pi(s)$ (actor) and a value function $V(s)$ (critic). This algorithm utilizes multiple learners that interact with their environment at the same time. The ``Advantage'' portion of the A3C algorithm consists of estimating how much better an action was than expected, calculated as $A = R - V(s)$. The ``Asynchronous'' portion of A3C comes from the fact that each learner is running individually in parallel. This dramatically speeds up learning. Each learner is initialized to the global policy. They then interact with each individual environment and calculate the value and loss. This is then used to calculate the gradients for the policy (Equation \ref{eqn:A3C_updatePolicy}) and value function (Equation \ref{eqn:A3C_updateValue}).

\begin{equation}
  d \theta \leftarrow d \theta + \triangledown_{\theta'} log \pi (a_i | s_i ; \theta' ) (R - V(s_i ; \theta'_v ))
  \label{eqn:A3C_updatePolicy}
\end{equation}
\begin{equation}
  d \theta_v \leftarrow d \theta_v + \partial (R - V(s_i ; \theta'_v ))^2 / \partial \theta'_v
  \label{eqn:A3C_updateValue}
\end{equation}

Where $\theta$ and $\theta_v$ are the parameters of the global policy and value function, and $\theta'$ and $\theta'_v$ are the parameters of the thread specific individual policy and value function. These are used to asynchronously update the global $\theta$ and $\theta_v$. 

DQN and A3C are two state of the art deep RL methods. The aim of this section was to give an overview of some recent developments in RL and is by no means a comprehensive review of RL as this is outside of the scope of this paper. For further reading on the state of the art in RL see the work of Arulkumaran et al. \cite{arulkumaran2017brief}. The next section will give a comprehensive account of the applications of RL to building energy management systems.

\section{Autonomous Building Energy Management via Reinforcement Learning}
\label{sec:BuildingManageRL}

An overview of both the problem of building energy management and RL was provided in Sections \ref{sec:smartBuildings} and \ref{sec:RL}. This section will now provide a comprehensive review of the literature that combines these two areas of research.

\subsection{HVAC}
\label{sec:RL_HVAC}

Reinforcement learning has been successfully applied to many areas of building energy management, including HVAC control. Typically, the environment states for the RL algorithm include factors such as: time of day, outdoor temperature, indoor temperature, weather forecast and occupancy. Typical RL actions include: temperature set points, air flow control, heating control and cooling control. The RL algorithm also needs some sort of reward to operate, all previous studies in the literature calculate rewards based on energy cost, thermal comfort or a combination of both.

The first application in the literature of RL to HVAC control was in 1996. Anderson et al. applied Q learning in conjunction with a PI controller to modulate the output of the PI controller for a heating coil \cite{anderson1996reinforcement,anderson1997synthesis}. This approach performed better than the PI controller alone. A 2006 study applied a combined RL - Model Predictive Control (MPC) approach to control HVAC operation at the Energy Resource Station Laboratory building in Ankeny, Iowa \cite{liu2006experimental}. It was found that the combined approach performed better than either Q learning or MPC individually. An actor critic - neural network learning approach was applied to adjust the signal of a local control for HVAC control in 2008 by Du and Fei \cite{du2008two}. This study reports significant improvements from a combined PID actor critic learning approach than a stand-alone PID controller. Dalamagkidis and Kolokotsa implemented the Recursive Least-Squares Temporal Difference (RLS TD) for HVAC control in 2008 and reported better performance with an RL controlled HVAC system than with a fuzzy PD controller \cite{dalamagkidis2008reinforcement}. Yu and Dexter implemented a Q($\lambda$) learning approach with fuzzy discretization of the state space variables to select rules that determine the operation of the HVAC system \cite{yu2010online}. A 2013 study by Urieli and Stone apply Tree Search to control a HVAC heat pump \cite{urieli2013learning}. The Tree Search method resulted in a 7 - 14\% saving when compared to the standard rule-based control. In 2015 Barrett and Linder applied Q learning to the problem of HVAC control combined with Bayesian Learning for occupancy prediction  \cite{barrett2015autonomous}. The results outline a 10\% energy savings improvement over a programmable control method. This work was then extended to apply parallel Q learning to HVAC control \cite{barrett2016automated}. Another 2015 study by Ruelens et al. implemented a combined Auto-Encoder (AE) Q learning to control a heat pump for a HVAC system \cite{ruelens2015learning}. The authors reported energy savings in the range of 4 - 11\%, comparable to the previous study.

Yang et al. implemented a Batch Q learning with a neural network to control a PV powered heating system \cite{yang2015reinforcement}. The results report a 10\% improvement on a standard rule-based system. In 2017 Wei et al. utilized deep neural network and applied Deep RL (DQN) to the problem of HVAC control and report energy saving improvements over conventional Q learning in the range of 20 - 70\% \cite{wei2017deep,wei2018model}. Another 2017 study by Wang et al. applied a Monte Carlo Actor Critic with Long Short Term Memory (LSTM) neural network to the task of HVAC control \cite{wang2017long}. Their results indicate a 15\% thermal comfort improvement and a 2.5\% energy efficiency improvement when compared to other methods. More recently, Marantos et al. proposed using a Neural Fitted Q-Iteration (NFQ) approach for HVAC control in 2018 \cite{marantos2018towards}. This approach also used a neural network to approximate the Q function. The results of this study report significant savings in terms of both energy consumption and thermal comfort when compared to rule-based controllers. In 2018, Zhang et al. applied Asynchronous Advantage Actor Critic (A3C) (a deep RL approach) to control the HVAC system for a simulation of the Intelligent Workplace building in Pittsburgh, USA \cite{zhang2018deep}. The authors report a 15\% energy saving improvement upon the base case. Chen et al. applied Q learning to control both the HVAC and window systems \cite{chen2018optimal}. The authors report energy savings of 13 and 23\% and lowered discomfort ratings by 62 and 80\% in the two buildings tested. In a 2018 study, Patyn et al. compared the performance difference of Q learning on three different neural network architectures (convolutional neural networks (CNN), LSTM networks and a feed forward multi-layer neural network) for HVAC control \cite{patyn2018comparing}. Their results indicate that the LSTM and multi-layer network perform best however the LSTM training time was twice as long.

This section has outlined numerous successful applications of RL to HVAC control with a variety of RL algorithms. The most common RL algorithm implemented was Q learning. The vast majority of studies report energy savings in the range of $\approx 10\%$ when compared to rule-based approaches. There is also a recent trend to apply deep RL to HVAC control, which has outperformed traditional tabular RL. This trend is unsurprising given the recent success and attention that deep learning has received.

\subsection{Water Heater}
\label{sec:RL_Water}

As discussed in Section \ref{sec:smartBuildings}, water heaters consume a significant amount of energy. This section will explore some of the applications of RL to control water heaters with the aim of reducing energy costs. Some of the state variables include: time of day, current water temperature and forecast usage. The action that the RL agent makes is generally to turn the heater on or off. The reward given to the agent is the electricity consumption.

A 2014 study by Al-Jabery et al. applied Q learning and a fuzzy Q learning to control a water heater \cite{al2014novel}. Their results indicate that their proposed fuzzy Q learning algorithm gives smoother convergence than the standard Q learning. Al-Jabery et al. then explored applying an Actor Critic - Q learning approach to control a water heater in 2017, where they included the grid load in the state space for the RL agent \cite{al2017demand}. This study reports energy cost savings between 6 and 26\% using RL. In 2014 Ruelens et al. applied batch Q learning to control a cluster of 100 water heaters \cite{ruelens2014demand}. The authors reported that within 45 days, the RL algorithm reduced electricity costs when compared to a hysteresis controller. In a more recent 2018 study, Ruelens et al. applied Q iteration to a physical water heater in a laboratory setting and found that Q iteration was able to reduce the energy consumption by 15\% over 40 days when compared to a thermostat controller \cite{ruelens2018reinforcement}. De Somer et al. applied Q learning to control the heating cycle of a domestic water heater to maximize the consumption from a local PV source. The authors then evaluate the Q learning agent on 6 residential houses and report a 20\% increase in the amount of local PV consumption \cite{de2017using}. Kazmi et al. did a similar study in the Netherlands where deep RL was applied to 32 homes \cite{kazmi2018gigawatt}. The authors report a 20\% reduction in the energy consumption for water heating and no loss in user comfort.

Although there are significantly fewer studies in the literature applying RL to water heaters than HVAC. Based on the papers reviewed, there appears to be a significant reduction in energy consumption when applying RL to water heaters. Many of the studies report reductions of $\approx 20\%$ when compared to the baseline. A similar trend can be observed that recent research is more focused on the application of deep RL as it has demonstrated its effectiveness.

\subsection{Home Management Systems}
\label{sec:RL_Home}

The studies outlined in Sections \ref{sec:RL_HVAC} and \ref{sec:RL_Water} consider applying RL to HVAC and water heaters respectively in isolation. This section explores applications of RL within the home to manage multiple appliances, lighting, PV, battery, etc. Applying RL in this manner is needed as the task of building energy management is a complex problem with multiple factors that must be managed to reduce the overall energy consumption. Similar to the previously mentioned sections, the states for the RL algorithm when exploring building energy management in a more holistic manner, generally consist of the time of day, temperature information and the current usage state of the various appliances. Many of these studies also include other state information such as electricity prices, grid load and information in relation to PV panels such as solar irradiance. The actions available to the RL agent in the studies outlined in this section are similar to those outlined in the previous sections, i.e. turning a device or appliance on or off. In the cases of studies which incorporate batteries, the actions are to charge/discharge the battery or to do nothing.

In 2018, a study by Reymond et al. applied Fitted Q Iteration to learn to schedule a number of household appliances including a heat pump, water heater and dishwasher \cite{reymondreinforcement}. Their results indicate that independent learning performs 9.65\% better than a centralized learning approach. Wei et al. implemented a dual iterative Q Learning algorithm for residential battery management \cite{wei2015novel}. The authors report a 32.16\% savings in energy cost when compared to the baseline. A 2015 study by Guan et al. utilize temporal difference learning to control the energy storage of a battery in the presence of a PV panel \cite{guan2015reinforcement}. It was found that temporal difference learning resulted in improvements of 59.8\% reduction in energy costs. Wan et al. implemented actor critic learning with two deep neural networks as a residential energy management system \cite{wan2018residential}. In this study, the RL algorithm learned when to purchase electricity from the utility provider based on previous prices, house hold loads and current battery charge. The authors report an 11.38\% reduction in energy cost by using actor critic learning when evaluated over 100 days. Remani et al. applied Q learning to schedule multiple devices within the home including lighting, clothes dryer, dish washer, etc. \cite{remani2018residential}. The authors implemented a price based demand response and included a PV panel in the system model and reported an approximate daily energy cost saving of 15\%. Wen et al. proposed a demand response energy management systems for small buildings that enables automated device scheduling in response to electrical price fluctuations \cite{wen2015optimal}. The authors implement Q learning with the aim of taking advantage of the estimated 65\% of potential energy savings for small buildings by efficient device scheduling and report improvements upon the baseline. Bazenkov and Goubko utilize inverse reinforcement learning (IRL) to predict consumer appliance usage and report a higher accuracy using IRL than other machine learning methods such as random forest \cite{bazenkov2018advanced}. A 2018 study by Mocanu et al. implement both Deep Q Learning (DQL) and Deep Policy Gradients (DPG) to optimize the energy management system for 10, 20 and 48 houses \cite{mocanu2018line}. This study explored the use of electric vehicles, PV panels and building appliances. The authors report electricity cost savings of 27.4 \% for DPG and 14.1\% for DQL.

A combined approach was proposed by Wang et al. that implements Q learning to manage a battery in a residential setting with installed PV \cite{wang2016near}. Q learning was combined with a model-based control algorithm to minimize energy costs. This hybrid approach demonstrated significant cost savings in the range of 23 - 72\%. A similar study was conducted by Mbuwir et al. that implemented fitted Q iteration for battery management in order to maximize the energy provided by PV \cite{mbuwir2017battery}. This was simulated using Belgian consumer data and demonstrated a cost saving of 19\%. Rayati et al. utilize Q learning for home energy management in a setting with PV installation and energy storage \cite{rayati2015optimising}. This study also considers home comfort and CO2 emissions when learning the optimal control policy. The authors report a maximum energy saving of 40\%, peak load reduction of 17\% and CO2 social cost reduction of 50\%. Sheikhi et al. implement Q learning to control a smart energy hub that consists of a combined heat and power, auxiliary boiler, electricity storage and heating storage \cite{sheikhi2016dynamic}. The authors report a saving of 30\% and 50\% for energy cost and peak loads respectively.

The results outlined in the literature relating to the application of RL to home energy management varies significantly more than in the previous two sections. The highest reported energy cost reduction was 72\% when implementing RL with a model-based controller \cite{wang2016near}. There is a greater capacity to generate savings when applying RL to home energy management systems with batteries and solar than simply applying RL to devices. This is reflected in the studies mentioned here. In terms of algorithms, Q learning is again the most routinely used method.

\subsection{Smart Homes and the Electrical Grid}
\label{sec:RL_Grid}

While the studies in the previous sections have demonstrated the effectiveness of RL when applied within the home, many studies in the literature examine how RL can also have a positive impact when considering how these smart homes are incorporated into the grid. This section will outline the applications of RL that relate to smart homes and the grid. Many of these studies evaluate control systems for a community of homes, rather than single homes in isolation, and therefore utilize multi-agent RL in many cases. There are many studies that focus on the application of RL to the operation of a microgrid to address problems such as thermal generator scheduling and incorporating PV into a microgrid. These are outside of the scope of this paper and will therefore not be discussed. This section solely focuses on studies that explore incorporating smart homes into the electrical grid. For a more general review of smart grid technologies, a comprehensive review is provided by Tuballa et al. \cite{tuballa2016review}.

Jiang et al. implemented a hierarchical multi-agent Q learning approach for dynamic demand response and distributed energy resource management in a microgrid \cite{jiang2015smart}. This microgrid also includes wind power and a battery. This study reports a 19\% reduction in energy costs for the whole community. Kim et al. implemented a multi-agent Q learning approach for residential energy consumption based on electricity prices in a microgrid \cite{kim2016dynamic}. Significant savings were reported over the baseline myopic purchasing strategy. A multi-agent RL approach was implemented by Anvari-Moghaddam et al. to integrate intelligent homes into the grid where batteries and renewables are also considered \cite{anvari2017multi}. The authors' proposed structure consisted of: a central coordinator agent, a building management agent, a renewables agent, a battery agent and a services agent. The authors report an energy cost saving of 5\%. A recent 2018 paper by Prasad and Dusparic implement a multi-agent deep RL approach to enable homes to share energy with one another in order to minimize cost for the community as a whole \cite{prasad2018multi}. The authors implemented a multi-agent DQN approach where each agent controlled a house. The agent could consume/store energy, request energy from or grant energy to neighbours, deny energy to neighbours or purchase from the grid. Each home was had an installed battery. The results indicate savings of 97 and 156kWh during the winter and summer respectively for a 10 home community.

Lu et al. recently implemented Q learning for electricity pricing with the aim to schedule pricing to maximize the providers profit when purchasing from an electricity wholesaler and minimize customers cost \cite{lu2018dynamic}. A 2018 study by Kofinas et al. implemented fuzzy Q learning to control the operation of a battery and desalination plant in conjunction with a PV panel for a microgrid that also includes residential electricity consumers \cite{kofinas2018energy}. Claessens et al. applied Q learning with a convolutional neural network for residential load control \cite{claessens2018convolutional}. It was found that the proposed approach provided near optimum performance when compared to a theoretically optimal benchmark. In 2018, Kim and Lim utilized Q learning to learn when to charge and discharge a battery and when to buy and sell from the grid \cite{kim2018reinforcement}. The authors report significant energy cost savings when compared to other methods.

The studies outlined in this section focus on how RL can reduce energy costs for multiple smart homes and how RL can help integrate these homes into the grid. As was observed in the previous sections, the studies here also demonstrate the effectiveness of RL for these problems. Multi-agent RL is a commonly used approach to learn policies for multiple homes as it is well suited to the distributed nature of this control task. Based on these studies, electricity cost savings can range from 5\% - 19\%. In terms of utilizing RL to balance residential loads with the grid, it was demonstrated that RL performs with near optimum performance and significantly reduces electricity costs for the grid.

\section{Discussion}
\label{sec:discussion}

A common theme across all of the various applications of RL within building energy management systems is that RL can provide significant savings in each problem it is applied to. As new and effective RL algorithms are developed, these new methods gradually make their way into the smart homes literature, e.g. deep RL methods. This section will outline some of the limitations of RL for building energy management and also some directions for future research.

\subsection{Limitations}
\label{sec:limitations}

There are however limitations of applying RL to building energy management systems. The overwhelming majority of studies outlined in this paper discuss applications of RL to simulated versions of problems relating to building energy management. While this is perfectly acceptable and a natural way to apply RL to these problems, this approach relies heavily on accurate simulator design and data that is representative of real-world scenarios. Since RL is an online learning algorithm, it could be directly applied to a building energy management system for a physical building without ever learning within a simulated environment. The problem with this approach is that in order for the RL algorithm to learn an effective control policy that minimizes energy cost, it must learn by trial and error. This would result in an initial exploration period where the RL agent would evaluate different policies, many of which would have an unacceptably high energy cost associated with it. Accurate simulators are needed for this reason, so that the RL agent can learn in simulation which policies are best. These pre-trained agents can then further refine their policies when implemented in physical systems. This is a viable approach as was demonstrated by the application of RL to the HVAC operation at the Energy Resource Station Laboratory building in Ankeny, Iowa \cite{liu2006experimental}.

\subsection{Future Directions}
\label{sec:future}
The first and most obvious trend that will be experienced relating to the application of RL to building energy management is the move towards deep RL algorithms. These deep RL methods are capable of learning more complex policies that are more sophisticated than those represented by shallow neural network or look up tables. As the volume of data collected by sensors will only increase, deep RL methods will become necessary to develop effective policies when interacting in environments with very large state action spaces. It is worth noting here that increasingly powerful computers will also enable researchers to train RL agents with increasingly complex policies.

Another potential route for future research in RL for building energy management is to apply many of the variants of traditional RL, some of which were outlined in Section \ref{sec:Problem}. For example, based on our extensive literature review, there are no examples within the literature that apply multi task RL to building energy management. Multi-task learning could be potentially useful for learning different HVAC control policies depending on how many occupants are in the house. Transfer learning could be useful for transferring knowledge learnt when learning to schedule appliances to scheduling water heaters.

Some of the problems relating to building energy management can be framed as multi-objective RL problems. The most obvious example of this is the trade of between thermal comfort and energy cost. There are numerous examples in the literature that apply multi-objective optimization algorithms to address this problem, however none of these studies utilize any state of the art multi-objective RL algorithms, e.g. Pareto Q learning \cite{van2014multi}.

An additional avenue for future research would be the application of meta learning to RL problems in building energy management, e.g. the RL\textsuperscript{2} algorithm mentioned in Section \ref{sec:Problem} \cite{duan2016rl}. There are no applications of meta reinforcement learning to building energy management. Methods such as RL\textsuperscript{2} could be useful for learning more effective control policies for the various within the studies outlined in this paper.

In terms of the tasks that RL is applied to, many studies in the literature are beginning to explore implementing multi-agent RL to control a community of homes rather than a single home in isolation. Developing control policies for groups of homes has the added benefit of allowing homes to pool their resources in a manner that results in greater savings for everyone. A natural route for this research to follow into the future would be to apply multi-agent RL to larger communities of homes to explore the scalability of multi-agent RL in this domain. Conducting game theory analysis in this area would also be a promising area of research. Decentralized multi-agent control policies for large communities of homes would rely heavily on cooperation to achieve cost savings for all, it would therefore be worthwhile to explore the issue of cooperation and defection in such systems.

Based on the extensive literature review conducted, all of the current research relating to applications of RL in building energy management evaluates systems in which the environment does not significantly change outside of its normal environmental conditions. This raises the question of how well do these learned control policies perform when faced with significant environmental changes, e.g. extreme weather events, battery degradation or failure, PV panal installation or failure, homes where the number of occupants increases or decreases, etc. It is not clear based on the current research, how the learned control policies can adapt to these significant environmental changes, i.e. are the learned policies brittle and would result in catastrophic failure with sub optimal performance in the event of these changes? If the RL agent is robust and capable of adapting to such environmental changes, how long would it take for the RL agent to relearn new policies in such scenarios? These questions would be worth investigating in depth in future studies.

\section{Conclusion}

This paper has discussed the key ideas behind developing building energy management systems, given a brief introduction into reinforcement learning and provided a comprehensive account of the applications of reinforcement learning to building energy management. As a result of this literature review, it is clear that:

\begin{enumerate}
    \item Reinforcement learning algorithms significantly improve the energy efficiency of homes. Energy savings vary significantly depending on the specific application. Broadly speaking RL can typically provide savings of $\approx 10\%$ for HVAC applications, $\approx 20\%$ for water heaters and $>20\%$ for more complete buildings energy management systems.
    
    \item The vast majority of studies that implement RL for problems relating to building energy management are in simulation only. With accurate simulator design however, RL is a viable approach for building energy management outside of simulation.
    
    \item Much of the recent research for RL and building energy management is focused on applications on deep RL algorithms due to their increased effectiveness over traditional approaches. This is likely to become more pronounced into the future.
    
\end{enumerate}

\bibliographystyle{unsrt}  


\end{document}